\documentclass{article}

\usepackage{arxiv}

\usepackage[utf8]{inputenc} % allow utf-8 input
\usepackage[T1]{fontenc}    % use 8-bit T1 fonts
\usepackage{hyperref}       % hyperlinks
\usepackage{url}            % simple URL typesetting
\usepackage{booktabs}       % professional-quality tables
\usepackage{amsfonts}       % blackboard math symbols
\usepackage{nicefrac}       % compact symbols for 1/2, etc.
\usepackage{microtype}      % microtypography
\usepackage{lipsum}
\usepackage{graphicx}
\usepackage{threeparttable}
\usepackage{multirow}
\graphicspath{ {./images/} }
\usepackage{algorithm}
\usepackage{algorithmic}
\usepackage{xcolor}
\usepackage{svg}
\usepackage{booktabs}
\usepackage{amssymb}
\usepackage{multirow}
\usepackage{siunitx}
\usepackage{threeparttable}
\usepackage{tablefootnote}

\usepackage{newfloat}
\usepackage{listings}

\floatstyle{ruled}
\newfloat{listing}{tb}{lst}{}
\floatname{listing}{Listing}

\title{EasySize: Elastic Analog Circuit Sizing via LLM-Guided Heuristic Search}

\author{
 Xinyue Wu\\
  Global College\\
  Shanghai Jiao Tong University\\
  800 Dong Chuan Rd., Shanghai, China \\
  \texttt{wuxinyue@sjtu.edu.cn} \\
   \And
 Fan Hu\\
  Global College\\
  Shanghai Jiao Tong University\\
  800 Dong Chuan Rd., Shanghai, China \\
  \texttt{hu-fan@sjtu.edu.cn} \\
  \And
 Shaik Jani Babu\\
  Global College\\
  Shanghai Jiao Tong University\\
  800 Dong Chuan Rd., Shanghai, China \\
  \texttt{skjanibabu786@sjtu.edu.cn} \\
\And
 Yi Zhao\\
  Global College\\
  Shanghai Jiao Tong University\\
  800 Dong Chuan Rd., Shanghai, China \\
  \texttt{joey-sjtu@sjtu.edu.cn} \\
\And
 Xinfei Guo\footnote{Corresponding author.}\\
  Global College\\
  Shanghai Jiao Tong University\\
  800 Dong Chuan Rd., Shanghai, China \\
  \texttt{xinfei.guo@sjtu.edu.cn} \\
}

\begin{document}
\maketitle
\begin{abstract}
Analog circuit design is a time-consuming, experience-driven task in chip development. Despite advances in AI, developing universal, fast, and stable gate sizing methods for analog circuits remains a significant challenge. Recent approaches combine Large Language Models (LLMs) with heuristic search techniques to enhance generalizability, but they often depend on large model sizes and lack portability across different technology nodes. To overcome these limitations, we propose EasySize, the first lightweight gate sizing framework based on a finetuned Qwen3-8B model, designed for universal applicability across process nodes, design specifications, and circuit topologies. EasySize exploits the varying Ease of Attainability (EOA) of performance metrics to dynamically construct task-specific loss functions, enabling efficient heuristic search through global Differential Evolution (DE) and local Particle Swarm Optimization (PSO) within a feedback-enhanced flow. Although finetuned solely on 350nm node data, EasySize achieves strong performance on 5 operational amplifier (Op-Amp) netlists across 180nm, 45nm, and 22nm technology nodes without additional targeted training, and outperforms AutoCkt, a widely-used Reinforcement Learning based sizing framework, on 86.67\% of tasks with more than 96.67\% of simulation resources reduction. We argue that EasySize can significantly reduce the reliance on human expertise and computational resources in gate sizing, thereby accelerating and simplifying the analog circuit design process. EasySize will be open-sourced at a later date.
\end{abstract}

% keywords can be removed
%\keywords{First keyword \and Second keyword \and More}

\section{Introduction}
Analog circuits are essential components of modern integrated circuits (IC)s, efficiently processing continuous signals with high precision, low power, and advanced capabilities.~\cite{Allen_Holberg_2012}. Their application scenarios include, but are not limited to audio/video processing, wireless communication, medical devices, and industrial automation.

Traditional analog circuits have fixed topologies, yet performing transistor gate sizing on them is not a straightforward task. The key challenges include: (1) \textit{High dependency on transistor technology nodes and topological structure.} Circuits at different nodes or with varying structures require unique parameters to meet design targets, making each sizing task distinct and hard to address by a universal automation algorithm. (2) \textit{Heavy reliance on human expertise.} Due to component nonlinearity, parasitic effect and the limited accuracy of theoretical calculations, the nominal parameters often require manual tuning based on extensive experiences from iterative simulations. (3) \textit{Multi-objective optimization requirements.} Take Op-Amp sizing, one of the most basic yet representative sizing tasks as an example, optimization is often expected to balance multiple conflicting objectives, such as gain ($A_{V}$), bandwidth (BW), phase margin (PM), bias direct current ($I_{DC}$) and slew rate (SR). As a result, some Reinforcement Learning (RL)-based methods, such as AutoCkt~\cite{settaluri2020autockt} require meticulous adjustment of initial points based on empirical experience, and may face the risk of overfitting. Besides, searching algorithms like Bayesian Optimization (BO)~\cite{pelikan2005bayesian}, Differential Evolution (DE)~\cite{lampinen2004differential} and Partical Swarm Optimization (PSO)~\cite{kennedy1995particle} used in knowledge-free sizing task, for example, Sizer~\cite{sizer}, still face severe instability in runtime and success rates.

\begin{figure}[!tbp]
\centering
\includegraphics[width=0.95\linewidth]{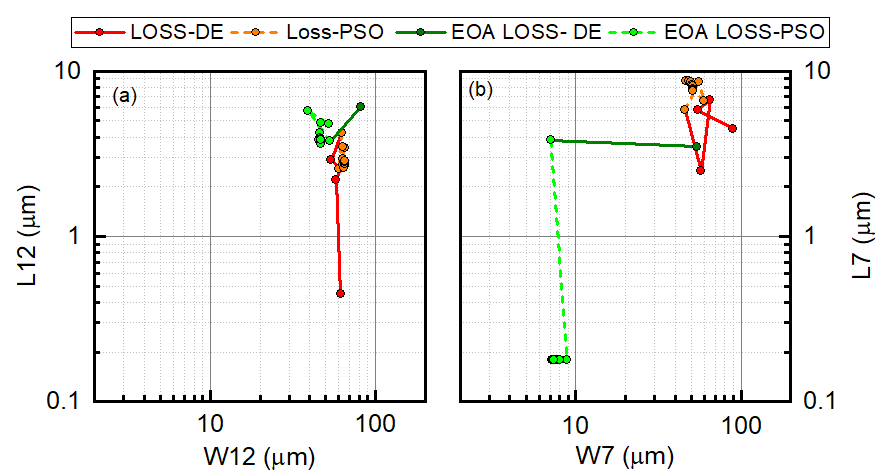}
\caption{\textbf{Best points movements of heuristic search under default and EOA-based modified loss functions.} (a) Influence on start point location. (b) Influence on movement orientation. }
\label{movement_track}
\end{figure}

The rise of Large Language Models (LLMs) has introduced new opportunities for automating analog circuit sizing. Existing works mainly focus on two aspects: (1) training-free prompt engineering agents like ADO-LLM~\cite{yin2024ado}, Atelier~\cite{shen2025atelier} and ChipMind~\cite{firouzi2025chipmnd}; (2) finetuned LLM frameworks~\cite{chen2024artisan}. However in both approaches, many LLMs directly output device sizes or downscaled search space for users, leading to the fact that it often involves extremely large models (70B parameters or more) to achieve sufficient accuracy. Besides, they are still heavily tied with certain process nodes or limited topologies, and are often tested with easy-to-achieve design tasks each with hours of runtime.

\begin{table}[!htbp]
\centering
\begin{threeparttable}
\begin{tabular}{cccccc}
\toprule
\textbf{Method} & \textbf{Light}\textsuperscript{1}& \textbf{Fast}\textsuperscript{2} & \textbf{Stable}\textsuperscript{3} & \textbf{General}\textsuperscript{4}&\textbf{Prior-free}\textsuperscript{5} \\
\midrule
Knowledge &   \checkmark          &  \(\times\)     &   \checkmark   & \(\times\) & \(\times\)      \\
gm/Id &  \checkmark     & \(\times\)       &  \checkmark        &   \(\times\) & \(\times\)       \\
Sizer &   \checkmark     &  \checkmark     &  \(\times\)       &   \checkmark   &   \checkmark    \\
AutoCkt &  \checkmark  &  \checkmark     & \checkmark       & \(\times\) & \(\times\)         \\
ADO-LLM &    \(\times\)         &  \checkmark     & \(\times\)       &   \checkmark  &\checkmark    \\
\textbf{EasySize}      &    \checkmark           &  \checkmark       &  \checkmark        &  \checkmark  &\checkmark       \\
\bottomrule

\end{tabular}
\caption{\textbf{Comparison of existing works.} Meanings of the columns: \textsuperscript{1}No extensive params. \textsuperscript{2}Hours finishable. \textsuperscript{3}Consistent circuit quality. \textsuperscript{4}Apply to new netlists or nodes. \textsuperscript{5}Without prior tuning experience. }
\label{tab:comparison_of_works}
\end{threeparttable}
\end{table}

To overcome these limitations and fully exploit the potential of LLMs in analog circuit sizing tasks, we introduce EasySize, a lightweight and stable LLM-enhanced searching framework that enables automatic analog circuit sizing within a short time. A comparison with previous methods is provided in Table~\ref{tab:comparison_of_works}. EasySize, for each run, dynamically generates loss functions to guide the search algorithms (DE combined with PSO), allowing more precise convergence toward satisfactory solutions. To further increase the agility, a feedback enhanced flow is integrated to react promptly to the current optimization state and refine previously generated loss functions.
% \begin{figure}[!tbp]
% \centering
% \includegraphics[width=0.95\columnwidth]{LaTeX/figs/Fig1_Search.png}
% \caption{\textbf{Best points movements of heuristic search under default and EOA-based modified loss functions.} (a) Influence on start point location. (b) Influence on movement orientation. }
% \label{movement_track}
% \end{figure}

In this work, we prioritize the Ease of Attainability (EOA) of different metric requirements in finetuning steps, ensuring the strong applicability even across nodes and topologies with significant variations. As illustrated in Fig.~\ref{movement_track}, the modified loss function with EOA can efficiently influence both start points and movement directions. Experiments demonstrate that EasySize, fine-tuned solely on a 350nm dataset, surpasses BO and AutoCkt in success rates and circuit quality metrics of most cases across 180nm, 45nm, and 22nm nodes for five distinct Op-Amp netlists. Additionally, EasySize significantly reduces simulations at 96.67\% compared to AutoCkt and at more than 10\% of 180nm and 45nm nodes compared to pure heuristic sizing.

%Experiments show that EasySize, fine-tuned solely on a 350nm dataset and even with simulation cost less than 3.33\% of AutoCkt's, outperforms BO under similar time constraints and AutoCkt at most of success rates and circuit quality metrics across 180nm, 45nm, and 22nm nodes for five different circuit netlists. It can also reduce simulations for two relatively easy-to-search technology nodes compared with pure heuristic sizing, therefore greatly saving computational resources and enhancing efficiency.

This paper makes four key contributions: 
\begin{itemize}
\item We present EasySize, the \textit{first} lightweight solution for analog circuit sizing that generalizes across both process nodes and circuit topologies, enabling high adaptability to diverse design scenarios.
\item EasySize introduces the \textit{first} attempt to leverage an LLM for generating loss functions that account for challenging metrics in gate sizing, revealing new opportunities for LLM integration in electronic design automation (EDA).
\item We propose a feedback-enhanced flow for dynamic loss adjustment using the LLM, enabling reflective optimization and improving the overall performance of the sizing process.
\item We develop the \textit{first} Ease of Attainability (EOA)-based benchmark generation method derived from simulation results. This approach maintains stable performance across various nodes and designs, providing a foundation for future zero-shot LLM applications in EDA.
\end{itemize}

% \begin{itemize}
% \item We introduce EasySize, the \textit{first} lightweight solution
% for both process node and topology generalizable Op-Amp sizing. It leads to significant adaptability to various scenarios.
% \item EasySize includes the \textit{first} try to generate loss function considering challenging metrics with LLM for size searching, exploring new potential for LLM integration in electronic design automation(EDA) area.

\section{Preliminary}
\label{sec:headings}

\subsection{Analog Circuit Sizing.}
In analog integrated circuit design, sizing methodologies play a crucial role in meeting performance specifications.  The square-law model, one of the earliest methodologies, presumes long-channel MOSFET behavior and employs a quadratic relationship between the drain current and gate-overdrive voltage. Nevertheless, it neglects short-channel effects and device parasitics in modern CMOS technologies, hence constraining its practical applicability~\cite{Allen_Holberg_2012}.  To overcome these restrictions, the gm/Id methodology employs transconductance efficiency (gm/Id) as a principal normalized design metric. This method offers enhanced flexibility and durability although it necessitates pre-generated lookup tables from simulations and thus susceptible to process variations~\cite{Jespers_Murmann_2017}.  As a result, hybrid methodologies that integrate these methods with targeted SPICE simulations are becoming more prevalent~\cite{Binkley_2007}. In recent years, there are also works using heuristic algorithms inside the design space for a fully knowledge-free blind search solution.

\subsection{Different EOA for Metrics in Search Space.}

\begin{figure}[!htbp]
    \centering
    \includegraphics[width=0.5\columnwidth]{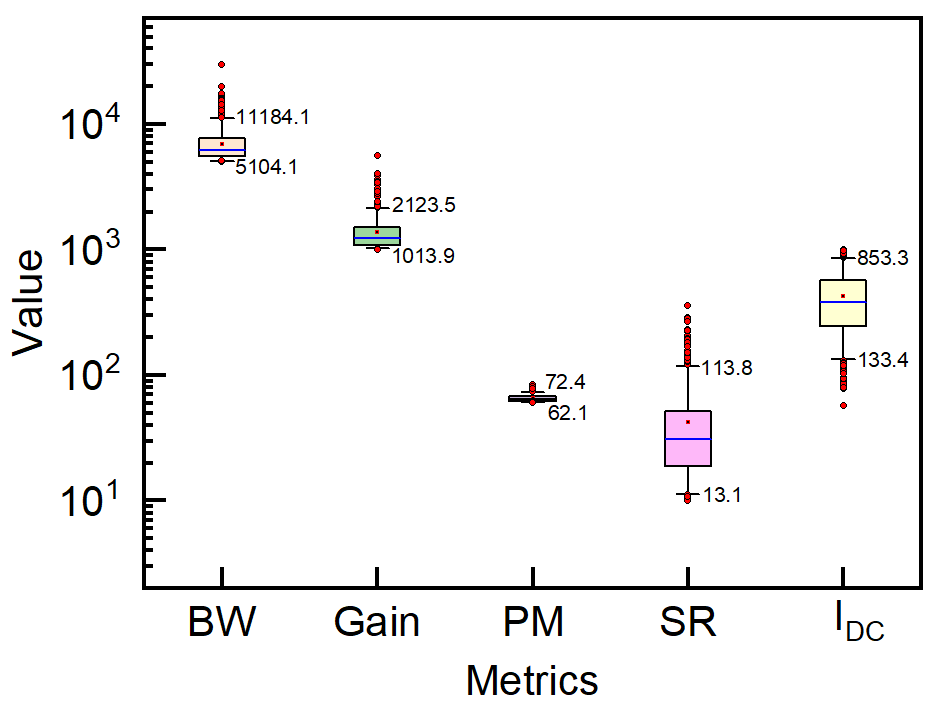}
    \caption{\textbf{The statistical box plot of five metric values over multiple simulations.} Units are: BW ($Hz$), Gain, PM ($^\circ$), SR ($V/\mu S$), $I_{DC}$ ($\mu A$). The units in the whole paper are all the same.}
    \label{fig:EOA_scatter_boxplot}
\end{figure}

Analog circuit sizing is a multi-objective optimization task with specific requirements. For Op-Amp sizing, they typically include gain ($A_V$), BW, PM, SR and $I_{DC}$. However, the feasibility of achieving these targets metrics varies significantly. As shown in Fig.~\ref{fig:EOA_scatter_boxplot}, based on 671 simulation results for the BSIM3V3.1 350nm~\cite{BSIM_350nm} A020005 netlist~\cite{sizer}, increasing BW from 5000 to 6000 is relatively achievable, whereas only a few design points reach a gain of 5000. This disparity arises not only from differences in metric units but also from the underlying physical properties. For example, the gain depends on $g_m$, which is directly proportional to the transistor $W/L$ ratio. While increasing $W$ and decreasing $L$ can boost gain, they also exacerbate parasitic capacitive loads at high frequencies, thus limiting BW. Similar trade-offs exist between gain and PM, SR and $I_{DC}$, as well as BW and $I_{DC}$, etc. Consequently, simultaneous extreme optimization of all metrics is unattainable in circuit design.

To analyze the achievability relationship, we propose the concept of Ease of Attainability (EOA) to indicate how easily a specific metric value can be achieved within a given set of design requirements. For a rough numerical estimation, EOA can be computed as the weighted average of the metric ranking $rank_i$ across all simulation results $S$:
\begin{equation}
rank_i = \frac{insert\_index(sort(S_i),metric_i)}{len(S_i)}
\label{formula:rank_calculation}
\end{equation}
\begin{equation}
EOA_{i} = \frac{rank_i}{\sum_{i=1}^{n} rank_i}
\label{formula:EOA_calculation}
\end{equation}

A lower EOA implies that the corresponding metric is more attainable under the same design conditions. However, EOA inherently depends on the predefined requirements before running simulations. Additionally, empirical rules influence the sizing process somehow. Therefore, EOA should be regarded only as a relative reference rather than an absolute metric, as it cannot be calculated with high precision. While different technology nodes and circuit topologies may slightly affect EOA values, these variations are usually limited. Hence, we adopt EOA as the key concept for designing a sizing loss function with strong versatility.

%However, EOA depends on the requirement before running simulation itself. Besides, there are some empirical rules during simulation, like PM usually has to be the first to guarantee, or it will be really hard to find a satisfiying point. Thus, EOA can only serve as a relative reference value and cannot be precisely calculated. Different technodes and topologies will slightly affect EOA, but often not too much, so we choose EOA as the breakthrough key for designing a sizing loss function with strong versatility.

\subsection{Heuristic Searching Algorithms}
Heuristic search is a kind of problem-solving strategy that employs practical methods and make informed guesses to find satisfactory solutions within a reasonable timeframe. Classical algorithms includes BO~\cite{pelikan2005bayesian}, DE~\cite{lampinen2004differential} and PSO~\cite{kennedy1995particle}, etc. BO uses probabilistic models to balance exploration based on uncertainty estimates. However, BO's reliance on prior distributions hampers rapid convergence, leading to massive computation with low success rates even for simple tasks. Therefore, it can only simulate about 100 points when DE or PSO has already found a solution. In contrast, DE iteratively improves candidate solutions through mutation, crossover, and selection operations based on differences among individuals in the population. It escapes local optima but often takes long to find a suitable solution due to random mutation directions. PSO updates particles' positions based on individual and swarm best experiences to find the suitable solution. It enables really quick search yet often get stuck locally in cases with complex dimensions.

Considering the characteristics of the aforementioned algorithms, we employ a combination of DE and PSO as the search component in this paper. To compare the circumstance of similar runtime, BO with 100 iterations (BO-100) serves as one of the baselines. 

\begin{figure*}[!t]
\centering
\includegraphics[width=1\linewidth]{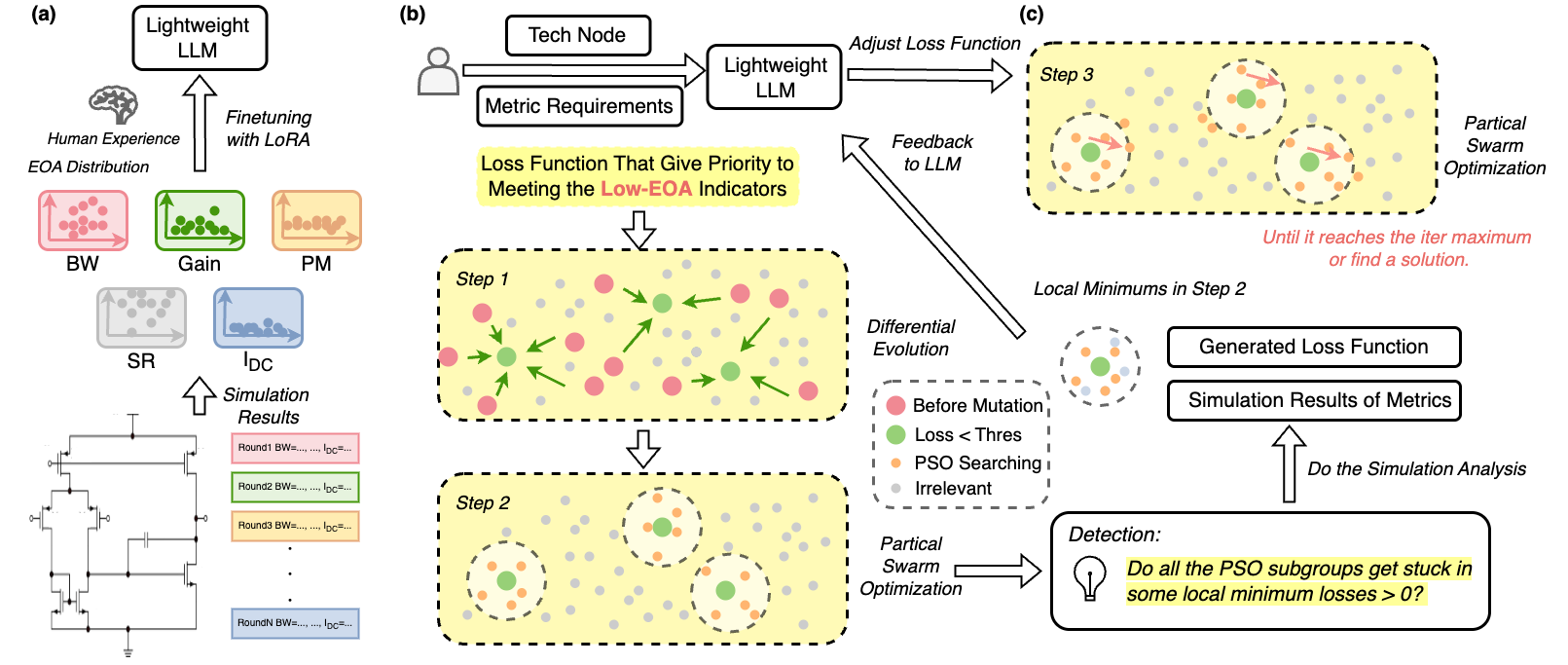}
\caption{\textbf{Method overview.} (a) The EOA dataset making and LoRA finetuning for LLM. (b) The first round of heuristic exploration based on DE and PSO. (c) The feedback enhanced step for dynamic loss readjustment. }
\label{fig_flow}
\end{figure*}
\section{Proposed Methodology}

\subsection{Methodology Overview}
We build EasySize from Sizer~\cite{sizer}, an open-source project for purely heuristic analog sizing. The overall flow of EasySize is explained in Fig.~\ref{fig_flow}.  First, the original Qwen3-8B~\cite{qwen3technicalreport} is fine-tuned with Low-Rank Adaptation (LoRA)~\cite{hulora} to generate the loss function according to various EOA of metrics. Next, DE performs a global search to identify promising parameter ranges, followed by PSO for local refinement. For each candidate design point, PySpice~\cite{pyspice} invokes an Ngspice~\cite{ngspice} subprocess to run simulations and compute the corresponding loss value. If all PSO processes stagnate and reach the maximum iteration limit, the simulation data of the best points are fed back to the LLM, which dynamically adjusts the previously generated loss function to enhance the search process.

% searches the possible good ranges according to the loss, then switches to PSO for local search. For each traversed point, PySpice calls an Ngspice subprocess to perform the simulation and calculate the loss value. If all PSO processes get stucked and reaches the maximum iteration limit, the current simulation information of the best points will be send back to LLM as feedback to adjust the previous loss.

\subsection{Dataset Preparation and LoRA Finetuning}
\begin{figure}[!h]
    \centering
    \includegraphics[width=0.6\linewidth]{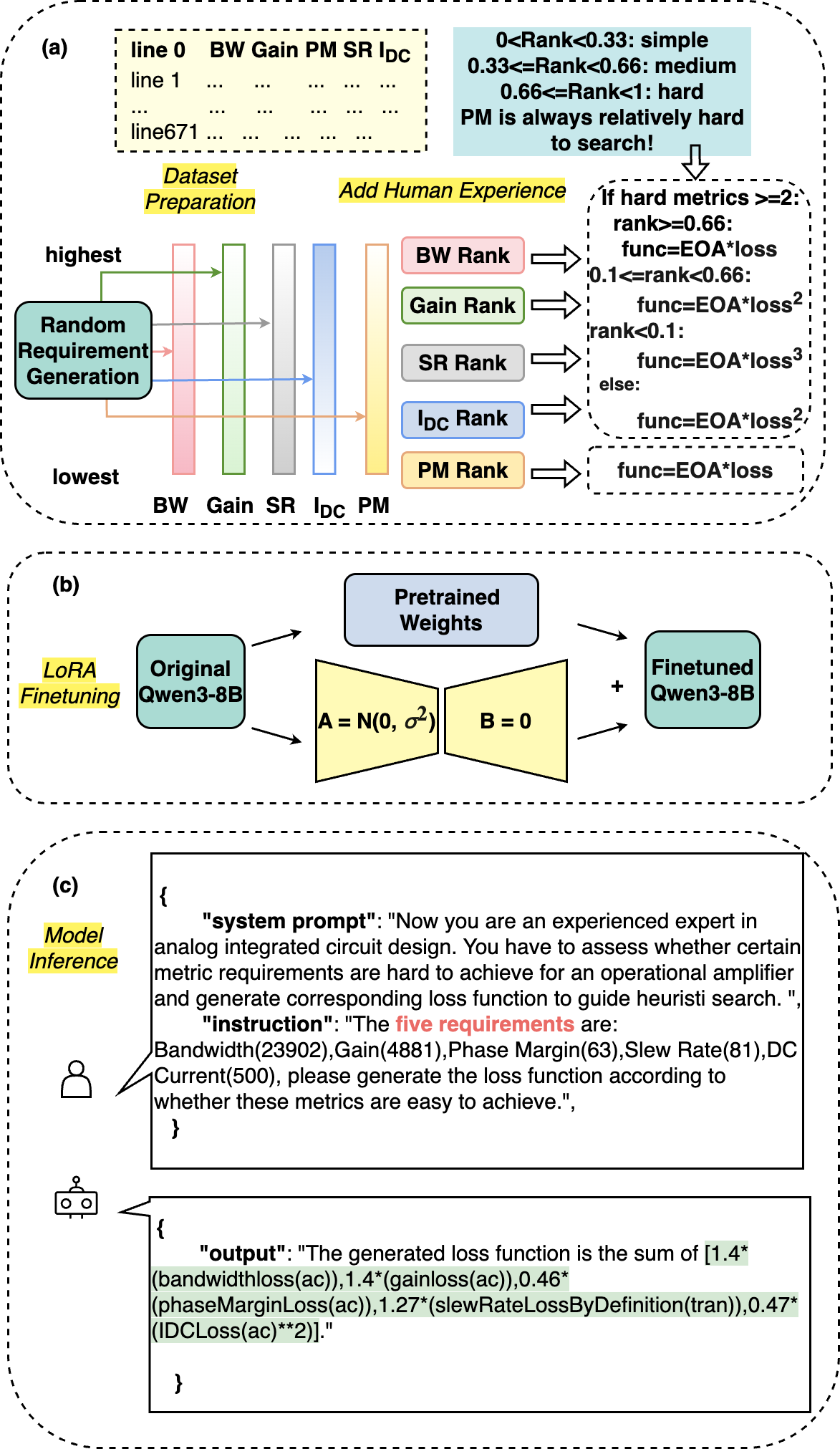}
    \caption{\textbf{Steps of preparation before the flow.} (a) Dataset generation according to EOA. (b) LoRA finetuning process. (c) Prompt and response for inference stage.}
    \label{fig:loss_finetune}
\end{figure}

Fig.~\ref{fig:loss_finetune} shows the finetuning and inference methods to enable LLM to gain knowledge of metric EOA values and loss function design. 

The dataset preparation method is shown in Fig~\ref{fig:loss_finetune}(a). To estimate EOA, we first conduct 671 DE search processes for five target metrics using a modified version of the A020005 netlist at the 350nm node provided in Sizer. To evaluate the relative difficulty of achieving baseline metrics, we set requirements for this initial round: BW $>$ 5000$Hz$, Gain $>$ 1000, PM in [60$^\circ$, 90$^\circ$] , SR $>$ 10$V/\mu S$, and $I_{DC}$ $<$ 1000$\mu A$. Based on the resulting data, we randomly generate multiple requirement settings within these ranges and calculate their corresponding ranks and EOAs using Formula~\ref{formula:rank_calculation} and Formula~\ref{formula:EOA_calculation}.

% The dataset preparation method is shown in Fig~\ref{fig:loss_finetune}(a). To obtain the EOA estimation value, we first run 671 DE searching processes of the five required metrics all with a 350nm A020005 modification version of netlist contains in Sizer. To assess the difficulty level of the baseline metrics, requirements for this round are set slightly more lenient as BW $>$ 5000, Gain $>$ 1000, PM in [60,90], SR $>$ 1 and $I_{DC}$ $<$ 1000. Then according to these values, we randomly generate different setting requirements inside their ranges, and calculate the corresponding ranks and EOAs based on Formula~\ref{formula:rank_calculation} and ~\ref{formula:EOA_calculation}. 

\begin{figure*}[!t]
\centering
\includegraphics[width=\linewidth]{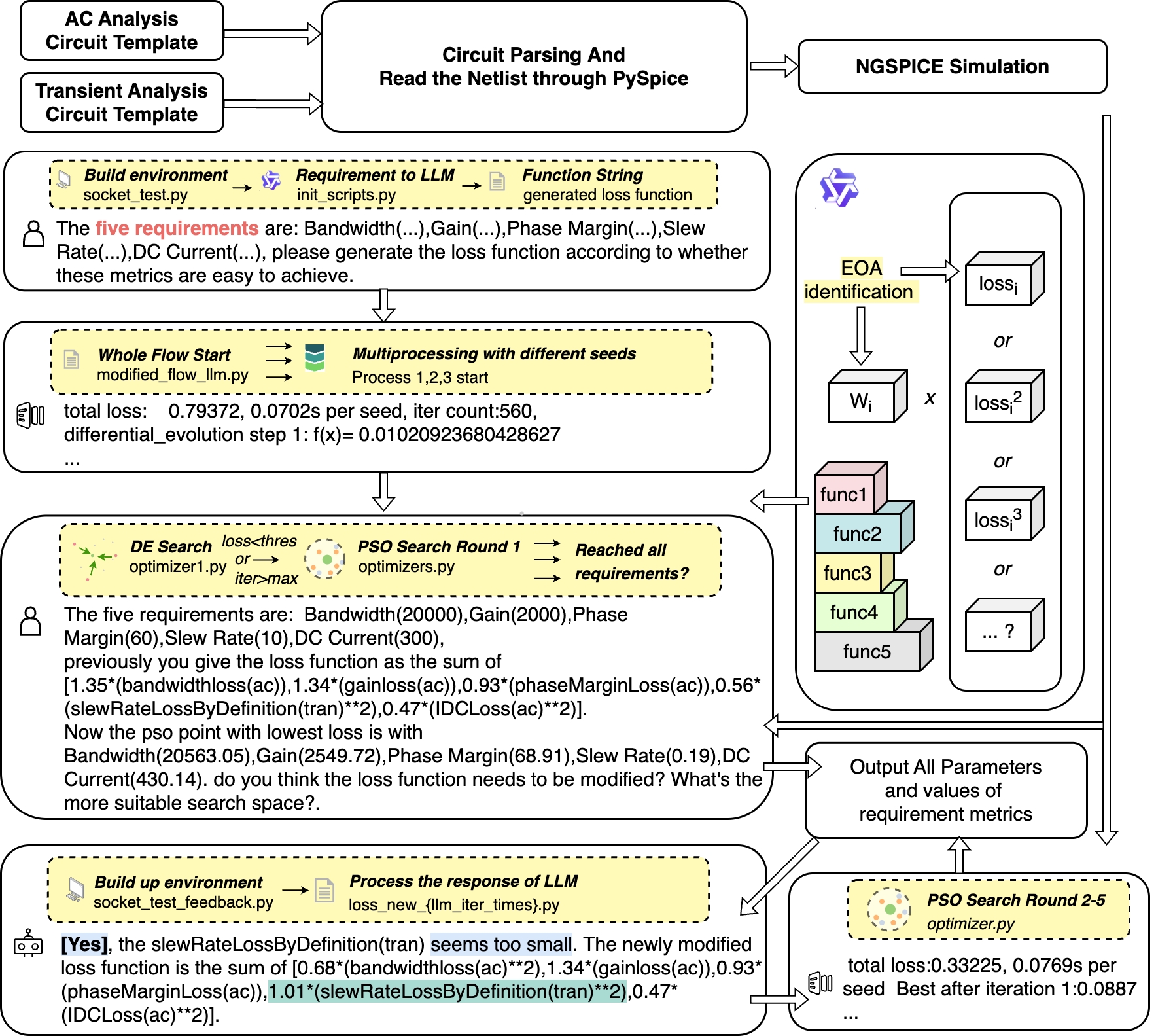}
\caption{\textbf{The feedback enhanced flow.}}
\label{fig_feedback}
\end{figure*}

Next, sample loss functions are obtained through the combination of EOA and human experience. There is a basic linear loss $loss_i$ for each $metric_i$:
\begin{equation}
loss_i = abs(\frac{min(w_i (metric_i - thres_i),0)}{thres_i})
\label{formula:basic_loss_format}
\end{equation}
In Formula~\ref{formula:basic_loss_format}, $w_{Gain} = w_{BW} = w_{SR} = 1 $, $w_{IDC} = -1$, with their thresholds the same as requirements. About PM, since its requirement is a range, we set $w_{PM}=1$ and $thres_{PM}=60^\circ$ for PM $<=$ 90$^\circ$,  and $w_{PM} = -1$ and $thres_{PM}=90^\circ$ for PM $>$ 90$^\circ$. However, this basic formulation is too simplistic to capture more complex cases. To address this, we classify the metrics into three difficulty levels—simple, medium, and hard—based on their ranks. Since based on experience, PM must be satisfied first, its loss always takes higher priority and is computed as the product of its EOA and linear loss. For scenarios where two or more metrics are classified as hard to achieve, it is often necessary to adjust the loss function format of the remaining four metrics. In these cases, the loss for each metric is calculated as its EOA multiplied by a linear, quadratic, or cubic version of the basic loss depending on its rank. Metrics with higher exponents are typically easier to satisfy, but their loss values tend to be small and therefore cannot be applied extensively. In all other cases, the default loss format for these four metrics is quadratic.

% Thus, we roughly grade the metrics as three levels (simple, medium, hard) according to their ranks. Since we have to first satisfy PM, its loss always takes higher priority as the product of its EOA and linear loss. After that, if there are two or more metrics are hard to get, it's possibly a case that needs to adjust loss function format of the rest four metrics to meet the demand, then the loss will be EOA multiplies the linear, qualdratic or cubic format of basic loss according to their ranks. Terms with higher exponents are typically easier to achieve, but their values are often too small to be used massively. Else, the four formats of loss will be default selected as qualdratic.

The Qwen3-8B model~\cite{qwen3technicalreport} is finetuned locally with the generated datasets following the process illustrated in Fig.~\ref{fig:loss_finetune}(b). In the LoRA approach, the original weight matrices $W$ of the pre-trained model are decomposed into two smaller matrices $A$ and $B$ such that $W = W_0 + AB$, where $W_0$ is the original weight matrix, while $A$ and $B$ are learnable low-rank matrices~\cite{hulora}. The formats of finetuning and inference prompts and responses are all the same, as shown in Fig.~\ref{fig:loss_finetune}(c).

To ensure the LLM learns only from data patterns and prompt cues, we do not expose the explicit generation process of the sample losses to the model. Interestingly, we observe that the LLM does not simply memorize training data or directly compute EOAs. Instead, it demonstrates the ability to generate loss functions in a way that mimics human reasoning, even for requirement settings it has never seen before. The weights it assigns to corresponding terms are not always numerically aligned with the calculated EOA values, yet they still produce effective guidance for the optimization process.

%To make LLM only learn through data patterns and prompt hints, we don't reveal the generation process of sample losses to LLM. It has been observed that LLM does not merely fit the training data and compute their EOA. Instead, it's capable of generating loss in a manner that resembles human judgment in requirement settings it has not encountered before, while the weights assigned to the corresponding terms are not always perfectly align with numeric EOA values.

\subsection{Feedback Enhanced Flow}
The feedback enhanced flow is integrated in EasySize to ensure its agility. Just as Fig.~\ref{fig_feedback} shows, after the first round of loss generation is done, DE and PSO begins to perform global and local search respectively. If all three PSO processes fail to find a satisfactory solution, the metrics of the current best points, along with the previously generated loss, are fed back to the LLM. The LLM then evaluates whether the existing loss formulation remains appropriate and adjusts it by increasing the weights of metrics that exhibit larger deviations from the required targets. The framework subsequently resumes PSO-based searching using the updated loss until either a valid solution is identified or the maximum iteration limit is reached.

\begin{table*}[!t]
\centering
\resizebox{\textwidth}{!}{
\begin{tabular}{@{}cc|ccc|ccc|ccc@{}}
\toprule
                              &      & \multicolumn{3}{c|}{\textbf{180nm}} & \multicolumn{3}{c|}{\textbf{45nm}}       & \multicolumn{3}{c}{\textbf{22nm}}   \\
                              &      & BO-100 & AutoCkt  & EasySize              & BO-100 & AutoCkt       & EasySize              & BO-100 & AutoCkt  & EasySize              \\ \midrule
\multirow{4}{*}{\textbf{T1}}  & AST\( \downarrow \)  & \textbf{98}     & 176400   & 3042            & \textbf{100}    & 324400        & 4958            & \textbf{100}    & 322380   & 6359            \\
                              & ADSR & 6.7\%  & 33.3\%   & \textbf{100.0\%}         & 0.0\%  & 66.7\%        & \textbf{93.3\%}          & 0.0\%  & 46.7\%   & \textbf{86.7\%}          \\
                              & PFoM & -1.8   & -19.5    & \textbf{0.5}             & -7.2   & -7.0          & \textbf{0.0}             & -23.9  & -41.3    & \textbf{-0.6}            \\
                              & MS\( \downarrow \)   & 2.0    & 1.0      & \textbf{0.0}             & 1.7    & 0.4           & \textbf{0.1}             & 2.3    & 1.1      & \textbf{0.3}             \\\midrule
\multirow{4}{*}{\textbf{T2}}  & AST\( \downarrow \)  & \textbf{100}    & 176400   & 6968            & \textbf{100}    & 324400        & 8735            & \textbf{100}    & 322380   & 10549           \\
                              & ADSR & 0.0\%  & 6.7\%    & \textbf{66.7\%}          & 0.0\%  & 20.0\%        & \textbf{33.3\%}          & 0.0\%  & 0.0\%    & \textbf{40.0\%}          \\
                              & PFoM & -42.6  & -18.6    & \textbf{-0.1}            & -97.8  & -12.1         & \textbf{-6.3}            & -128.7 & -53.4    & \textbf{-11.6}           \\
                              & MS\( \downarrow \)   & 2.4    & 1.9      & \textbf{0.9}             & 2.1    & \textbf{1.7}           & \textbf{1.7}             & 2.9    & 2.3      & \textbf{1.5}             \\\midrule
\multirow{4}{*}{\textbf{T3}}  & AST\( \downarrow \)  & \textbf{100}    & 176400   & 3690            & \textbf{100}    & 324400        & 4077            & \textbf{100}    & 322380   & 8462            \\
                              & ADSR & 0.0\%  & 33.3\%   & \textbf{73.3\%}          & 0.0\%  & \textbf{73.3\%}        & 40.0\%          & 0.0\%  & \textbf{53.3\%}   & 33.3\%          \\
                              & PFoM & -23.6  & -18.7    & \textbf{-0.9}            & -420.8 & \textbf{0.6}           & -2.6            & -121.6 & -44.1    & \textbf{-6.6}            \\
                              & MS\( \downarrow \)   & 2.2    & 1.7      & \textbf{0.3}             & 1.9    & \textbf{0.3}           & 1.3             & 2.3    & \textbf{1.2}      & 1.6             \\\midrule
\multirow{4}{*}{\textbf{T4}}  & AST\( \downarrow \)  & \textbf{100}    & 176400   & 6774            & \textbf{100}    & 324400        & 8640            & \textbf{100}    & 322380   & 9055            \\
                              & ADSR & 0.0\%  & 0.0\%    & \textbf{66.7\%}          & 0.0\%  & 20.0\%        & \textbf{46.7\%}          & 0.0\%  & 7.0\%    & \textbf{53.3\%}          \\
                              & PFoM & -35.4  & -21.7    & \textbf{-0.5}            & -126.3 & \textbf{-0.8}          & -22.4           & -55.0  & -55.3    & \textbf{-2.1}            \\
                              & MS\( \downarrow \)   & 3.0    & 2.1      & \textbf{1.1}             & 2.6    & 1.1           & \textbf{1.0}             & 3.1    & 1.8      & \textbf{1.3}             \\\midrule
\multirow{4}{*}{\textbf{T5}}  & AST\( \downarrow \)  & \textbf{100}    & 176400   & 7607            & \textbf{100}    & 324400        & 9239            &\textbf{100}    & 322380   & 9406            \\
                              & ADSR & 0.0\%  & 0.0\%    & 0.0\%           & 0.0\%  & 0.0\%         & 0.0\%           & 0.0\%  & 0.0\%    & 0.0\%           \\
                              & PFoM & -100.7 & -24.2    & \textbf{-16.3}           & -144.3 & \textbf{-6.3}          & -36.9           & -263.8 & -58.0    & \textbf{-36.5}           \\
                              & MS\( \downarrow \)   & 2.8    & 3.5      & \textbf{2.4}             & 2.5    & 2.9           & \textbf{2.3}             & 2.4    & 3.1      & \textbf{2.3}             \\ \midrule
\multirow{4}{*}{\textbf{Avg}} & AST\( \downarrow \)  & \textbf{99.6}   & 176400.0 & 5616.0          & \textbf{100.0}  & 324400.0      & 7057.7          & \textbf{100.0}  & 322380.0 & 8766.1          \\
                              & ADSR & 1.3\%  & 16.0\%   & \textbf{61.3\%} & 0.0\%  & 36.0\%        & \textbf{42.7\%} & 0.0\%  & 21.3\%   & \textbf{42.7\%} \\
                              & PFoM & -40.8  & -20.6    & \textbf{-3.5}   & -159.3 & \textbf{-5.1} & -13.6           & -118.6 & -50.4    & \textbf{-11.5}  \\
                              & MS\( \downarrow \)   & 2.5    & 2.1      & \textbf{1.0}    & 2.2    & \textbf{1.3}  & \textbf{1.3}    & 2.6    & 1.9      & \textbf{1.4}    \\ \bottomrule
\end{tabular}
}
\caption{\textbf{Main results.} "\( \downarrow \)" refers to the-lower-the-better indicators.}
\label{tab:main_results}
\end{table*}

\section{Experiment}
We conduct extensive experiments to evaluate the capability of EasySize. Both model finetuning and framework execution are performed on an Ubuntu 20.04 system equipped with an NVIDIA GeForce RTX 4090 GPU. We set the number of DE candidate points to three and employ the same number of multi-subprocesses to improve runtime efficiency. To reduce randomness, in each method, the tasks corresponding to each combination of technology node and netlist are run 3 times separately to obtain average values.

\begin{table}[!b]
\centering
\begin{tabular}{cccccccc}
\toprule
\textbf{Task} & \textbf{BW($Hz$)}    & \textbf{Gain} & \textbf{PM($^\circ$)}    & \textbf{SR($V/\mu S$)} & \textbf{I$_{DC}$($\mu A$)} & \textbf{Level}  \\
\midrule
T1 & $>$5k  & $>$1k & 60-90 & -  & -   & easy \\
T2 & $>$20k & $>$1k & 60-90 & $>$3  & -   & mid \\
T3 & $>$5k  & $>$2k & 60-90 & $>$3  & -   & mid \\
T4 & $>$5k  & $>$1k & 60-90 & $>$10 & $<$300 & mid \\
T5 & $>$20k & $>$2k & 60-90 & $>$10 & $<$300 & hard \\  
\bottomrule
\end{tabular}
\caption{\textbf{Task settings of different requirements.}     }
\label{task_settings}
\end{table}
\subsection{Benchmark}
The evaluation benchmark is retrieved from Analog Designer's Toolbox (ADT) standard library designs~\cite{ADT_website, ADT_Hesham2023}. All Op-Amp circuits have been designed with BSIM PTM technology nodes namely 180nm, 45nm, and 22nm~\cite{ptm_BSIM}. The dissimilar process nodes and topology variations are selected to test the generalizability across scenarios in industry application. The five corresponding design tasks generated from a multitude of search results and human expertise are shown in Table~\ref{task_settings}, where ``Level'' indicates whether the parameters meeting the task can be easily searched. Among them, Task 1 is the most common and basic requirement, while Task 5 is nearly impossible to achieve.

\setlength{\tabcolsep}{1mm}
% Please add the following required packages to your document preamble:
% \usepackage{multirow}
\begin{table*}[!t]
\resizebox{\textwidth}{!}{
\begin{threeparttable}
\begin{tabular}{cc|cccc|cccc|cccc}
\toprule
                              &      & \multicolumn{4}{c|}{\textbf{180nm}}                            & \multicolumn{4}{c|}{\textbf{45nm}}         & \multicolumn{4}{c}{\textbf{22nm}}           \\
                              &      & ES    & ES/LLM & DE & PSO & ES              & ES/LLM & DE     & PSO    & ES              & ES/ LLM & DE     & PSO    \\ \midrule
\multirow{4}{*}{\textbf{T1}}  & AST \( \downarrow \) & 3042            & \textbf{1497}            & 3239        & 1887         & 4958            & 3621   & 5802   & \textbf{3120}   & 6359            & 3701    & 5695   & \textbf{3627}   \\
                              & ADSR & \textbf{100.0\%}         & \textbf{100.0\%}         & 80.0\%      & 73.3\%       & \textbf{93.3\%}          & 80.0\% & 40.0\% & 33.3\% & \textbf{86.7\%}          & 73.3\%  & 53.3\% & 33.3\% \\
                              & PFoM & 0.5             & \textbf{0.6}             & \textbf{0.6}         & 0.3          & 0.0             & \textbf{0.2}    & \textbf{0.2}    & -0.2   & -0.6            & -9.2    & \textbf{0.2}    & -0.2   \\
                              & MS \( \downarrow \)  & \textbf{0.0}             & \textbf{0.0}             & 0.2         & 0.8          & \textbf{0.1}             & 0.6    & 0.9    & 1.9    & \textbf{0.3}             & 0.7     & 1.0    & 2.0    \\
                              \midrule
\multirow{4}{*}{\textbf{T2}}  & AST\( \downarrow \)  & 6968            & 6537            & 7618        & \textbf{4233}         & 8735            & 9267   & 6923   & \textbf{3440}   & 10549           & 10300   & 6942   & \textbf{4480}   \\
                              & ADSR & \textbf{66.7\%}          & 53.3\%          & 46.7\%      & 13.3\%       & \textbf{33.3\%}          & \textbf{33.3\%} & 20.0\% & 0.0\%  & \textbf{40.0\%}          & 20.0\%  & 20.0\% & 0.0\%  \\
                              & PFoM & -0.1            & \textbf{0.0}             & -1.1        & -4.0         & \textbf{-6.3}            & -9.6   & -152.6 & -34.7  & -11.6           & -13.3   & -188.3 & \textbf{-4.5}   \\
                              & MS \( \downarrow \)  & \textbf{0.9}             & 1.3             & 1.3         & 2.5          & \textbf{1.7}             & 2.1    & 2.4    & 2.2    & \textbf{1.5}             & 2.7     & 2.1    & 3.1    \\
                              \midrule
\multirow{4}{*}{\textbf{T3}}  & AST \( \downarrow \) & \textbf{3690}            & 6717            & 6584        & 4280         & \textbf{4077}            & 8971   & 6942   & 4273   & 8462            & 9633    & 6942   & \textbf{4240}   \\
                              & ADSR & \textbf{73.3\%}          & 46.7\%          & 46.7\%      & 13.3\%       & \textbf{40.0\%}          & 6.7\%  & 20.0\% & 13.3\% & \textbf{33.3\%}          & 26.7\%  & 6.7\%  & 6.7\%  \\
                              & PFoM & -0.9            & -208.6          & \textbf{-0.1}        & -0.6         & \textbf{-2.6}            & -95.1  & -435.9 & -13.0  & -6.6            & \textbf{-1.9}    & -494.3 & -24.3  \\
                              & MS \( \downarrow \)  & \textbf{0.3}             & 1.5             & 1.3         & 2.6          & \textbf{1.3}             & 2.5    & 2.0    & 2.3    & \textbf{1.6}             & 2.1     & 2.2    & 2.9    \\\midrule
\multirow{4}{*}{\textbf{T4}}  & AST \( \downarrow \) & 6774            & 6318            & 5393        & \textbf{3407}         & 8640            & 8212   & 6656   & \textbf{3787}   & 9055            & 7627    & 6942   & \textbf{3900}   \\
                              & ADSR & \textbf{66.7\%}          & 46.7\%          & 46.7\%     & 33.3\%       & \textbf{46.7\%}          & 26.7\% & 20.0\% & 20.0\% & \textbf{53.3\%}          & 46.7\%  & 20.0\% & 20.0\% \\
                              & PFoM & -0.5            & -10.8           & \textbf{-0.4}        & -0.8         & \textbf{-22.4}           & -120.2 & -304.6 & -49.7  & \textbf{-2.1}            & -7.0    & -6.3   & -31.1  \\
                              & MS \( \downarrow \)  & \textbf{1.1}             & 1.6             & 1.6         & 2.4          & \textbf{1.0}             & 2.1    & 2.0    & 2.3    & \textbf{1.3}             & 2.2     & 2.5    & 2.7    \\\midrule
\multirow{4}{*}{\textbf{T5}}  & AST \( \downarrow \) & 7607            & 10320           & 6942        & \textbf{4507}         & 9239            & 10427   & 6942   & \textbf{4573}   & 9406            & 9102    & 6942   & \textbf{4627}   \\
                              & ADSR & 0.0\%           & 0.0\%           & 0.0\%       & 0.0\%        & 0.0\%           & 6.7\%  & 0.0\%  & 0.0\%  & 0.0\%           & 0.0\%   & 0.0\%  & 0.0\%  \\
                              & PFoM & \textbf{-16.3}           & -341.6          & -88.1       & -37.1        & \textbf{-36.9}           & -141.2 & -376.2 & -190.0 & \textbf{-36.5}           & -52.7   & -206.7 & -94.7  \\
                              & MS \( \downarrow \)  & \textbf{2.4}             & 3.7             & 2.8         & 3.5          & \textbf{2.3}             & 2.7    & 2.4    & 3.1    & \textbf{2.3}             & 3.3     & 3.1    & 3.3    \\ \midrule
\multirow{4}{*}{\textbf{Avg}} & AST \( \downarrow \) & 5616.0          & 6277.6          & 5954.1      & \textbf{3663.0}       & 7057.7          & 8099.8 & 6653.6 & \textbf{3839.0} & 8766.1          & 8072.7  & 6693.7 & \textbf{4175.0} \\
                              & ADSR & \textbf{61.3\%} & 49.3\%          & 44.0\%      & 26.7\%       & \textbf{42.7\%} & 30.7\% & 20.0\% & 18.7\% & \textbf{42.7\%} & 33.3\%  & 20.0\% & 12.0\% \\
                              & PFoM & \textbf{-3.5}   & -112.1          & -17.8       & -8.5         & \textbf{-13.6}  & -73.2  & -253.8 & -57.5  & \textbf{-11.5}  & -16.8   & -195.1 & -31.0  \\
                              & MS \( \downarrow \)
                              & \textbf{1.0}    & 1.6             & 1.4         & 2.4          & \textbf{1.3}    & 2.0    & 1.9    & 2.4    & \textbf{1.4}    & 2.2     & 2.2    & 2.8    \\ \bottomrule
\end{tabular}
% \begin{minipage}{}\footnotesize
% \textsuperscript{1} EasySize. 
% \textsuperscript{2} EasySize without LLM, which is a DE+PSO searching framework. 
% \end{minipage}
\caption{\textbf{Ablation results.} ES here stands for EasySize, and ES/LLM stands for EasySize without LLM, which is a DE+PSO searching framework. "\( \downarrow \)" refers to the-lower-the-better indicators.}
\label{tab:ablation}
\end{threeparttable}
}
\end{table*}

\subsection{Evaluation Metrics}
We use average simulation times (AST), all-demands-successful rate (ADSR), Penal Figure of Merits (PFoM) and Miss of Specs (MS) to measure the framework performance. AST represents the average number of Ngspice simulations required to complete a single task. ADSR denotes the percentage of runs that successfully meet all specified requirements. MS measures the number of performance metrics that fail to satisfy the design specifications.

For more rational circuit quality assessment, we define the PFoM based on Figure of Merits (FoM)~\cite{yin2024ado}, a metric widely used in gate sizing tasks. Traditional FoM often fails to capture extreme trade-offs; for instance, if one metric has an exceptionally high value while another is extremely low, the overall FoM may still appear acceptable, even though the circuit is non-functional. Thus, we introduce PFoM to penalize such extreme underperformances. The calculation of PFoM is given by the following formula:
\begin{equation}
    PFoM = \sum w_i \times \frac{val_{metric_i}-req_{metric_i}}{val_{metric_i}}
\end{equation}
This metric will impose greater penalties on those with significantly non-compliant indicators. For PFoM calculation, the weights are set as $w_{Gain} = w_{BW} = w{SR} = 1$, $w_{IDC} = -1$, $w_{PM} = 1$ while PM $<=$ 90$^\circ$, and for cases that PM $>$ 90$^\circ$, PM will be transformed to (150$^\circ$-PM) to measure. For values equaling 0, to prevent the PFoM from measuring excessively large values that could lead to unstable evaluation, the corresponding values will be set to: BW = 1$Hz$, Gain = 1, PM = 1$^\circ$, SR = 0.01$V/\mu S$. $I_{DC}$ is not applicable to this circumstance because it already satisfies the requirement when it's near 0.

\subsection{Main Results}
Table~\ref{tab:main_results} compares the efficiency and generation quality of our LLM-enhanced EasySize framework with traditional heuristic BO algorithm (implemented with Scikit-learn 1.5.1 ~\cite{scikit-learn}) and AutoCkt~\cite{settaluri2020autockt}, an RL-based analog sizer. The maximum iteration step limits of DE, PSO in EasySize and BO are set with empirical experience respectively as 25, 50 and 100, and if PSO iteration steps $>$ 10, it is regarded as "stuck". AutoCkt is trained to near convergence using a single set of trade-off specifications: BW = 5000$Hz$, gain = 2000, PM = 60$^\circ$, SR = 3 $V/\mu s$, and $I_{DC}$=300$\mu A$. However, training of AutoCkt failed on 180nm A020007 and A020009 netlists due to the failure of finding the initial point that meets the -3dB bandwidth requirement, and are inferenced with their 45nm checkpoints with AST denote as 0.

The result demonstrates that EasySize  achieves a high number of successful cases and superior circuit quality while maintaining a runtime of the same order of magnitude as BO-100. Furthermore, it outperforms AutoCkt on 86.67\% of total tasks with reduction of more than 96.67\% of latter’s simulation resources, proving its strong zero-shot generalizability. In contrast, AutoCkt performs the best on Task 3, the one closest to its training specifications, but it failed to generalize to different cases and even exhibits overfitting on the A020014 netlist at 45nm and 22nm nodes, resulting in a low ADSR for the simplest case.

\subsection{Ablations}
We evaluate the effectiveness of different components in EasySize through the results presented in Table~\ref{tab:ablation}. The configuration ``ES/LLM'' is tested with the default loss function, which is defined as the sum of quadratic terms with equal weights of 1 without the feedback flow. The configurations ``DE'' and ``PSO'' refer to pure heuristic flows using only Differential Evolution or Particle Swarm Optimization, respectively, without any LLM enhancement. The maximum iteration limits are kept consistent with those in the main experiments. The results show that the LLM component significantly improves the stability of the entire flow, increasing ADSR performance by an average of 30.33\% and reducing MS by 21.76\%. Moreover, it lowers respectively 10.54\% and 12.87\% of AST required for the two simpler technology nodes (180nm and 45nm), demonstrating its efficiency benefits.

%We assess the effectiveness of various components in EasySize with results in Table ~\ref{tab:ablation}. The flow of ``ES/LLM'' is run with the corresponding default loss function (the sum of corresponding quadratic functions with weights all equal to 1) and no feedback flow is integrated. "DE" and "PSO" indicate the two pure flows with only these algorithms without LLM enhancement. The maximum iteration step limits are the same as the main experiment. Results show that LLM part ensures the stability of the whole flow, averagely enhances 30.33\% of the ADSR performance, and decrease 21.76\% of MS. It also reduces some simulation needs for the two simpler technodes.

\section{Conclusion}
This paper proposes EasySize, a lightweight LLM-enhanced analog circuit sizing framework with strong generalizability across different process nodes and circuit topologies. The LLM is finetuned to generate appropriate loss functions based on the EOA of metric requirements, enabling faster and more effective heuristic search. Furthermore, EasySize integrates a feedback-enhanced design flow, which further strengthens its optimization performance and adaptability.

\section*{Acknowledgment}
This work was supported by the National Science and Technology Major Project (Grant No. 2021ZD0114701).

\bibliographystyle{unsrt}  
%\bibliography{references}  %%% Remove comment to use the external .bib file (using bibtex).
%%% and comment out the ``thebibliography'' section.

%%% Comment out this section when you \bibliography{references} is enabled.
\bibliography{references}

\end{document}